\documentclass{article}

\usepackage{arxiv}

\usepackage[utf8]{inputenc} 
\usepackage[T1]{fontenc}    
\usepackage{hyperref}       
\usepackage{url}            
\usepackage{booktabs}       
\usepackage{amsfonts}       
\usepackage{nicefrac}       
\usepackage{microtype}      
\usepackage{graphicx}
\usepackage{doi}
\usepackage{amsmath,amsfonts}
\usepackage{algorithmic}
\usepackage{algorithm}
\usepackage{array}
\usepackage{amssymb}
\usepackage{cleveref}       
\usepackage{bm}
\usepackage{siunitx}
\usepackage{multirow}
\usepackage{booktabs}

\title{Differentiable Collision-Supervised Tooth Arrangement Network with a Decoupling Perspective}

\date{}

\newif\ifuniqueAffiliation
\uniqueAffiliationfalse
\ifuniqueAffiliation 
\author{ Zhihui He\thanks{Use footnote for providing further
		information about author (webpage, alternative
		address)---\emph{not} for acknowledging funding agencies.} \\
	School of Software, Tsinghua University\\
	\texttt{hezh22@mails.tsinghua.edu.cn} \\
	\And
	{Chengyuan Wang\thanks{}} \\
	School of Software, Tsinghua University\\
	\texttt{chengyua22@mails.tsinghua.edu.cn} \\
	\AND
	Shidong Yang\\
	\texttt{email} \\
	\And
	Li Chen\\
	\\
	Address \\
	\texttt{email} \\
	\And
	Shuo Wang \\
	Affiliation \\
	Address \\
	\texttt{email} \\
	\And
	Yanheng Zhou \\
	Affiliation \\
	Address \\
	\texttt{email} \\
}
\else
\usepackage{authblk}

\newbox{\orcid}\sbox{\orcid}{\includegraphics[scale=0.06]{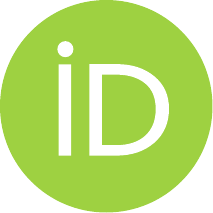}} 
\author[1]{%
	Zhihui He\thanks{Equal contributions.}%
}
\author[1]{%
	Chengyuan Wang\textsuperscript{*}%
}
\author[1]{%
	Shidong Yang%
}
\author[1]{%
	Li Chen\thanks{Corresponding author.}%
}
\author[2]{%
	Yanheng Zhou%
}
\author[3]{%
	Shuo Wang%
}

\affil[1]{School of Software, Tsinghua University}
\affil[2]{Tsinghua University Affiliated Beijing Tsinghua Changgung Hospital, Dental}
\affil[3]{Peking University Hospital of Stomatology, Orthodontic}

\affil[ ]{\{hezh22, chengyua22\}@mails.tsinghua.edu.cn}
\affil[ ]{chenlee@tsinghua.edu.cn}

\fi

\renewcommand{\shorttitle}{}
\hypersetup{
pdftitle={Differentiable Collision-Supervised Tooth Arrangement Network with a Decoupling Perspective},
pdfkeywords={3D Vision, Orthodontics, Tooth Arrangement, 6-DoF Pose Prediction.},
}

\begin{document}
\maketitle

\begin{abstract}
Tooth arrangement is an essential step in the digital orthodontic planning process. Existing learning-based methods use hidden teeth features to directly regress teeth motions, which couples target pose perception and motion regression. It could lead to poor perceptions of three-dimensional transformation. They also ignore the possible overlaps or gaps between teeth of predicted dentition, which is generally unacceptable. Therefore, we propose DTAN, a differentiable collision-supervised tooth arrangement network, decoupling predicting tasks and feature modeling. DTAN decouples the tooth arrangement task by first predicting the hidden features of the final teeth poses and then using them to assist in regressing the motions between the beginning and target teeth. To learn the hidden features better, DTAN also decouples the teeth-hidden features into geometric and positional features, which are further supervised by feature consistency constraints. Furthermore, we propose a novel differentiable collision loss function for point cloud data to constrain the related gestures between teeth, which can be easily extended to other 3D point cloud tasks. We propose an arch-width guided tooth arrangement network, named C-DTAN, to make the results controllable. We construct three different tooth arrangement datasets and achieve drastically improved performance on accuracy and speed compared with existing methods. 
\end{abstract}


\section{Introduction}

Orthodontics is the branch of dentistry concerned with facial growth, development of dentition and occlusion, and the diagnosis, interception, and treatment of occlusal anomalies \cite{littlewood2019introduction}. In the field of orthodontics, the integration of deep learning has not only provided new perspectives for the optimization of treatment strategies but has also significantly enhanced the precision and efficiency of diagnostics and therapeutic interventions\cite{wei2022dense,cui2021tsegnet,cui2022fully}. Tooth arrangement is an essential step in the digital orthodontic planning process. It aims to generate target aesthetical dentitions that are compliant with orthodontic rules by arranging three-dimensional teeth models. With these target dentitions, the aligner can be designed to force each tooth to move to its target pose.
\begin{figure}[h]
    \centering
    \includegraphics[scale=0.6]{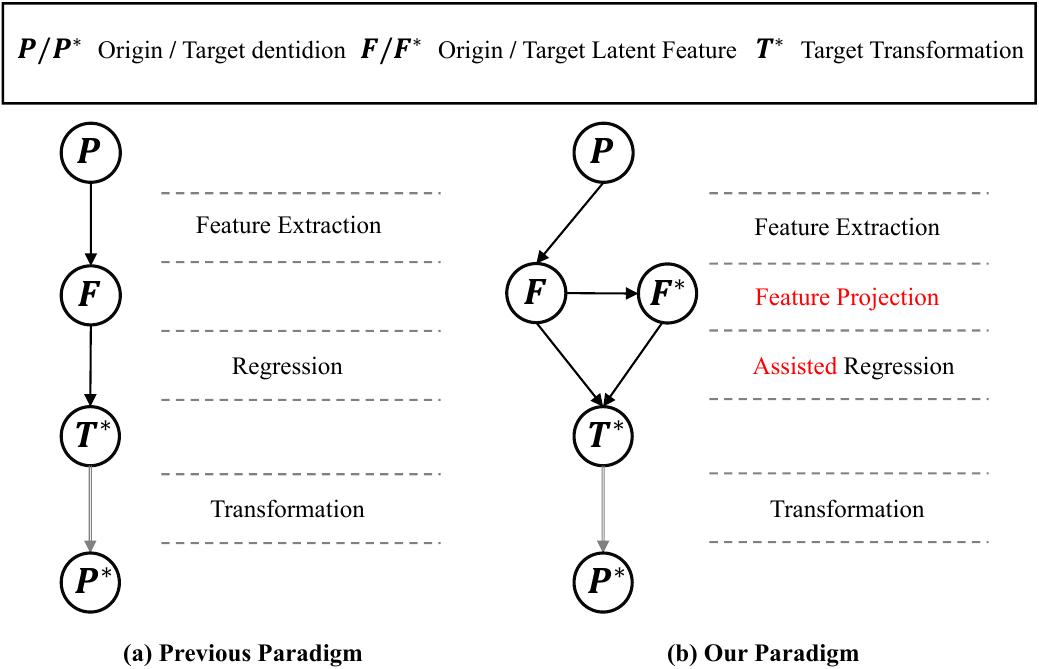}
    \caption{Comparisons on different paradigms of previous arrangement methods and ours.}
    \label{fig:paradigms}
\end{figure}
However, the current tooth arrangement process is mainly completed through manual interactions, and the process is time- and labor-consuming for the experts. Generally, it often takes a few days after doctors upload the related orthodontic materials. Moreover, based on the complexity of 3D object interactions, it is difficult for inexperienced technicians to generate satisfying arrangement results.
Using orthodontic rules to constrain the final dentition could be helpful to the automatic tooth arrangement task, but the rules are usually complicated and not comprehensive. Therefore, some learning-based methods \cite{li2020malocclusion,wei2020tanet,lingchen2020iorthopredictor,wang2022tooth} have been proposed to address this task. These methods aim to achieve the final position of the arranged teeth, with the corresponding point cloud data as the processing object. These methods have achieved impressive results, showcasing the enormous potential of learning-based methods. Nevertheless, there are still some certain gaps for actual clinical application. 1) The overall quality of tooth arrangement, both in functionality and aesthetics, remains a significant bottleneck. 2) Previous methods overlooked the necessity of ensuring adjacent teeth are correctly attached, a critical condition that must be met. 3) There is also a practical need to control the arch width of arrangement results in specific situations.

\begin{figure}
    \centering
    \includegraphics[scale=0.35]{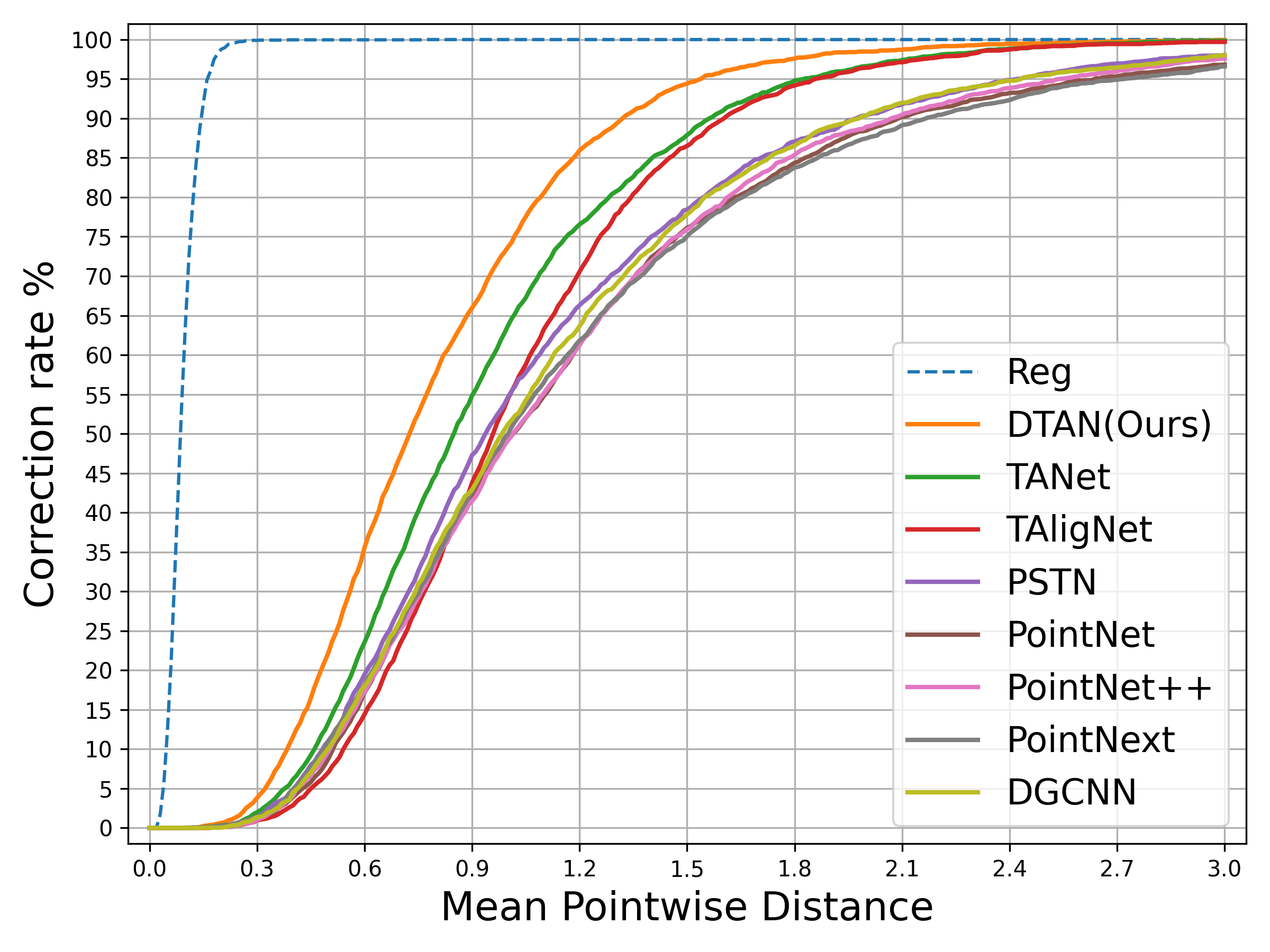}
    \caption{Comparison of the errors between our method and others. The abscissa represents the threshold of the mean pointwise distance (mm), and the lines depict the accuracy of each method under this threshold. "Reg" indicates that we use the location features extracted from the ground truth as $\bm F^*$ to assist the network in transform parameter regression, resulting in almost no error.}
    \label{fig:intro_vis}
\end{figure}

To address these issues, we first summarize the previous work paradigms. They all use an encoder (e.g., PointNet \cite{qi2017pointnet}) to extract point cloud features and a decoder (e.g., Multi-Layer Perceptron) to predict rigid transformation for each tooth to maintain geometric invariance. As shown in Figure \ref{fig:paradigms}(a), they get the final dentition through transformation parameters. However, regarding transformations as actions performed by experts, there must be target location perceptions before actions. So the previous paradigm actually couples location perceptions with transformation parameter regression. It is indirect and challenging for networks to learn the expression of the transformation parameters. In Figure \ref{fig:paradigms}(b), we demonstrate the paradigms of our method. Complying with the human experts' workflow, we decouple this task into target pose perception and assisted transformation parameter regression. Since it is difficult to perceive the target poses in 3D space, inspired by recent works\cite{bigalke2021fusing,zhu2017unpaired,Rombach_2022_CVPR}, we let the network perceive them at the feature-level with feature consistency supervision, making it easier for the network to express richer information. After obtaining the features of the original dentition $\bm F$, we first use the feature projector to perceive the positional features $\bm F^*$ of the final dentition, then concatenate $\bm F$ with $\bm F^*$ to regress the transformation parameters through the decoder. 
It is worth noting that, as Figure \ref{fig:intro_vis} shows, as long as an accurate perception of the final dentition is obtained, the error in the regression process (Reg) is negligible.

Furthermore, specific rules must be satisfied in the final dentition; for example, overlaps or gaps between teeth are not allowed. Unfortunately, previous works have not considered this constraint. To enforce this constraint during the learning process, we need a function simultaneously measuring the degree of embedding and spacing between two teeth. While there are methods to detect collisions, discrete binary supervision is challenging to optimize in the learning process. Spacing can be measured by Chamfer distance\cite{fan2024collaborative}, but embeddings cannot. There are also methods to calculate the signed distance between two teeth, which cannot be applied to learning-based tasks without computable gradients. Additionally, the attached surface between two occluding teeth may be non-convex. In our method, we propose a novel loss function that measures the depths of overlaps or the lengths of gaps. This loss function enables a differentiable and parallelizable learning process in point cloud data. Impressively, it can be easily extended to other 3D point cloud tasks.

There is also an actual demand to adapt to the target dentitions and diagnoses of patients. For instance, the teeth roots must remain within the range of the alveolar bone. Therefore, the results of tooth arrangement should be controllable. Previous methods did not achieve this functionality, as they could not produce reasonable tooth arrangement results under simple and easily adjustable conditions. In our method, we design a new network using the arch widths of the upper and lower jaws as conditions to generate diverse and controllable results.

Combining the above points, as shown in Figure \ref{fig:paradigms}, our method achieves the most minor error compared with other methods. Our contributions can be explained as follows:
\begin{itemize}
    \item We decouple the tooth arrangement task into feature-level target pose perception and assisted transformation regression tasks, which helps reduce the complexity of it. And we decouple teeth features into geometric features and positional features for each tooth with feature-level consistency supervision to ensure the effectiveness of feature-level target pose perception.
    \item We introduce an approximate differentiable collision loss function and validate its effectiveness in a variety of situations to avoid potential overlaps or gaps in the final dentition.
    \item We introduce a slight arch-width guided tooth arrangement network which is able to make the results diverse and controllable.
    \item We construct three different datasets to validate our method and get the best performance compared with other state-of-the-art methods. Moreover, we obtain over two-time inference speed-up increase by the previous SOTA methods \cite{wei2020tanet,wang2022tooth}.
\end{itemize}
Our code and some typical cases will be released soon.

\section{Related Work}

\subsection{Deep Architectures for Point Cloud Analysis}
Since point clouds are unordered and irregular, convolutional neural networks, which have had tremendous success in images, can not be applied directly. Considering the characteristic of permutation invariance, PointNet \cite{qi2017pointnet} uses MLP layers to extract point-wise features and a max-pooling layer to integrate the global feature. In order to better capture local geometric details of neighboring points, PointNet++ \cite{qi2017pointnet++} is proposed with a set abstraction layer. This proves that grouping operation is critical in capturing geometric details. Following this idea, some graph-based methods with different grouping operations are proposed. DGCNN \cite{wang2019dynamic} proposed EdgeConv group the nodes on dynamic graphs generated in the feature space. In addition, some Spatial CNN-based \cite{tatarchenko2018tangent,li2018pointcnn} use grids or pseudo grids to make it allow for convolution. Recently, some transformer-based methods \cite{lai2022stratified,yu2022point} use self-attention to capture global or local features. PointNext \cite{qian2022pointnext} revisits these previous methods and finds better architectures and training strategies on open datasets with a mount of data. Since the global information of the whole dentition is vital, these base models for point clouds are not suitable for directly applying to tooth arrangement. In our work, we choose PointNet\cite{qi2017pointnet} as our feature extraction backbone, and we are surprised to find that, with the simplest feature extraction backbone, our network achieves impressive results in both quality and convergence speed.

\subsection{Automatic Tooth Arrangement}

Some methods have already been proposed for data-driven automatic tooth arrangement over the years. Two contemporaneous works \cite{li2020malocclusion,wei2020tanet} both claim that they are the first learning-based work for tooth arrangement. The former one, PSTN \cite{li2020malocclusion}, utilizes the global features of the whole dentition and local features of each tooth to predict the transformation matrices, which could generate non-rigid deformations of teeth. The latter one, TANet \cite{wei2020tanet}, with the utilization of global features and local features as well, further proposes a graph-based Feature Propagation Module to model the relations between teeth. TAligNet \cite{lingchen2020iorthopredictor} is proposed as a part of facial image generation. A landmark-based method \cite{wang2022tooth} is proposed to enhance the relations between tooth and tooth by landmark points constraint. However, landmark labeling needs the experience and time of experts, which is challenging to obtain. The quality of the arrangement depends on the quality of the landmark, which is extremly demanding. All these methods above regression the transformation directly, which may be elusive for neural networks. They also ignore the necessary attaching constraints between teeth or need landmarks to benefit model learning. Unlike these methods, we propose a method to decouple the motion regression process and use approximate differentiable collision supervision to constrain teeth attaching. In addition, the results of tooth arrangements could be diverse as long as they are aesthetically pleasing and satisfy orthodontic rules. TANet \cite{wei2020tanet} uses a random conditional vector to generate different results, while these results are not controllable. However, the doctor will also customize a tooth arrangement plan based on the patient’s demands and factors such as his alveolar bone and facial shape. Therefore, we further design an arch-width guided tooth arrangement network to generate diverse and controllable results.

\begin{figure}
    \centering
    \includegraphics[scale=0.22]{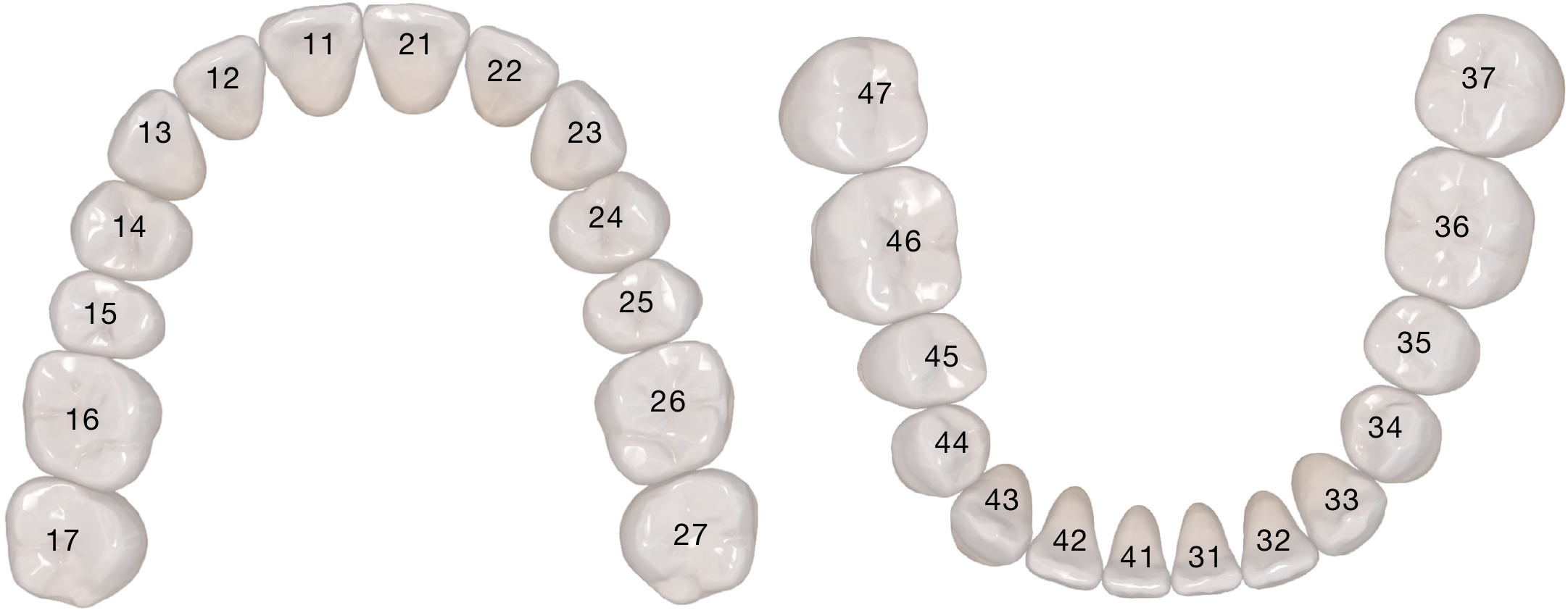}
    \caption{Input teeth meshes and corresponding teeth labels. The first digit indicates the quadrant where the teeth are located. The second digit indicates the classes of teeth: 1-2 for incisors, 3 for cuspids, 4-5 for bicuspids, and 6-7 for molars.}
    \label{fig:tooth_labels}
\end{figure}

\begin{figure*}[tb]
    \centering
    \includegraphics[scale=0.14]{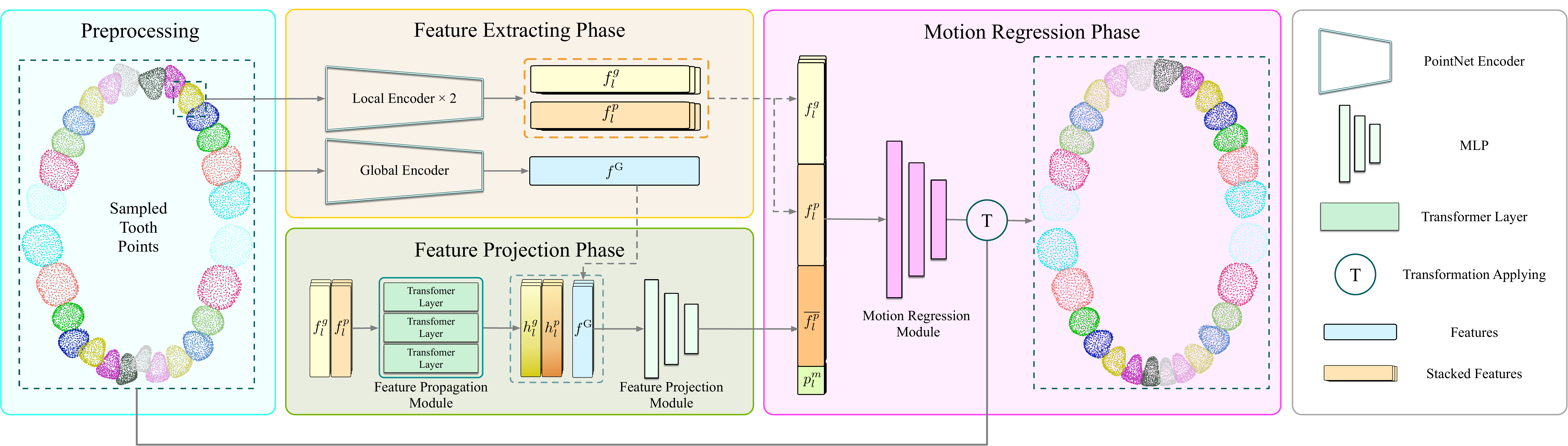}
    \caption{Overview of DTAN. We use a three-phase network to generate the transformation parameters of our arrangement. The first phase contains one global encoder and two local encoders to extract global features, local geometric features, and local positional features. The second phase propagates teeth features and then projects them to hidden features of target dentition. Then, the third phase integrates features and generates transformation parameters. At last, the parameters are applied to the original input for the final dentition result.}
    \label{fig:overview}
\end{figure*}

\section{Method}

\subsection{Overview}

The input data is a set of point clouds of labeled teeth $\bm{P} = \{\bm{P}_l \subseteq \mathbb{R}^{N\times 3} | l \in \bm{L} \}$ where $\bm{P}_l$ denotes the sampled points from tooth $l$, $\bm{L}$ is the set of teeth labels and $N$ is the number of points for each tooth. Figure \ref{fig:tooth_labels} demonstrates each tooth before sampling to point cloud and its label. The goal for tooth arrangement is finding a rigid transformation $\bm{T}_l$ for each tooth and transforming each tooth to get an aesthetically pleasing result. The formulaic representation is as follows:
\begin{equation}
    \min_{f} \sum_{\bm{P} \in \mathcal{P}}\sum_{l \in \bm{L}}||\overline{\bm{P}_l} - \bm{P}_l^*||^2
    \label{eq:goal}
\end{equation}
where $\mathcal{P}$ is the arrangement dataset. $\overline{\bm{P}_l}$ is the predicted teeth generated by predicted transformation $\bm{T}_l$ of deep model. And $\bm{P}_l^*$ indicates the ground truth after being transformed by ground truth transformation $\bm{T}_l^*$.

As shown in Figure \ref{fig:overview}, our \textbf{DTAN} takes point clouds of teeth as input and outputs 6-DoF transformations. The final dentition results can be calculated by applying the transformation parameters to the original point clouds. Our DTAN can be divided into three phases. The first phase is the feature extracting phase. In this phase, we extract the decoupled geometric, positional, and global features of the whole dentition by point cloud encoders. The second phase is the feature projecting phase, which projects the hidden features of the original dentition to the target dentition. The last phase, the motion regressing phase, regresses the 6-DoF arrangement motion using the original and the projected features.

\subsection{Details}

In this section, we introduce the details of the three phases in DTAN.
In the first phase, we use two point cloud feature extracting encoders $\phi^{\text{L}} = \{\phi^{\text{Geo}}, \phi^{\text{Pos}}\}$ based on PointNet to extract the hidden features of each tooth. Because teeth need to maintain the original position and geometric shape, we remove the spatial transformation network from it. There is a slight difference between the inputs of two encoders. $\phi^{\text{Pos}}$ uses point cloud of teeth $\bm{P}_l$ as input while $\phi^{\text{Geo}}$ uses $\widetilde{\bm{P}_l}$ as input. $\widetilde{\bm{P}_l} = \{\bm{p}' = \bm{p} - \bm{p}^m_l | \bm{p} \in \bm{P}_l, \bm{p}^m_l = \frac{\sum_{\bm{p}\in \bm{P}_l}\bm{p}}{|\bm{P}_l|}\}$, where $\bm{p}^m_l$ is the barycenter of tooth $\bm{P}_l$. The feature extracting processing can be described as follows:
\begin{equation}
    \bm{f}^g_l = \phi^{\text{Geo}}(\widetilde{\bm{P}_l}), ~\bm{f}^p_l = \phi^{\text{Pos}}(\bm{P}_l)
\end{equation}
where $\bm{f}^g_l, \bm{f}^p_l \in \mathbf{R}^C$ are local geometric features and positional features of tooth $l$ with dimension $C$.
To enhance the capture of global information, we add an additional global feature extractor $\mathcal{\phi}^{\text{G}}$ which is described as follows:
\begin{equation}
    \bm{f^{\text{G}}} = \phi^{\text{G}}(\bm{P}, \bm{C})
\end{equation}
where $\bm{C}$ is the set containing barycenters of teeth. The input for the global feature extractor consists of points and barycenters that belong to the points of each tooth.

The second phase contains two modules: the Feature Propagation Module and the Feature Projection Module. Recognizing the challenge for the global encoder to capture local geometric details, we add the Feature Propagation Module to propagate teeth features effectively. It enables the model to perform the tooth arrangement task using enhanced and more informative features. The process is shown as follows:
\begin{equation}
\begin{aligned}
    \bm{h}^g_l = \psi^{\text{Geo}}(\bm{f}^g_l),~
    \bm{h}^p_l = \psi^{\text{Pos}}(\bm{f}^p_l)
\end{aligned}
\end{equation}
where $\bm{h}^g_l, \bm{h}^p_l \in \mathbf{R}^C$ are the propagated geometric features and positional features of tooth $l$. $\psi^{\text{Geo}}$ and $\psi^{\text{Pos}}$ are both Feature Propagation Modules.
In DTAN, we use Vanilla Transformer Encoder \cite{vaswani2017attention} as our Feature Propagation Module. With the Attention Mechanism, the module can better capture the characteristic relationships between teeth, which helps predict the overall shape of the final dentition (e.g., the dental arch) and accelerates the convergence speed.
The next step is predicting each tooth's target features, which we call feature projection. While the geometric features of each tooth before and after arrangement are consistent, it is ordinary to predict the target positional features only. In this step, DTAN aggregates global features and local features to predict the result features as follows:
\begin{equation}
    \overline{\bm{f}^p_l} = \psi^{\text{Proj}}(\bm{f}^{\text{G}}, \bm{h}^g_l, \bm{h}^p_l).
\end{equation}
We choose MLP as the Feature Projection Module, and its output $\overline{\bm{f}^p_l}$ should have the same dimension with $\bm{f}^p_l$.

Finally, in the third phase, we regress the arrangement motions of each tooth. Referring to registration, the regressor $\omega^\text{Mo}$ predicts the motion parameters between the original and the target poses. In order to reduce the difficulty of regression, we further concatenate geometric features $\bm{f}^g_l$ and rotation centers, which is barycenter $\bm{p}^m_l$ in our settings, to positional features $\bm{f}^g_l, \bm{f}^p_l$ as the block input. The process is shown as follows:
\begin{equation}
    (\overline{\bm{t}_l}, \overline{\bm{q}'_l}) = \omega^\text{Mo}(\bm{f}^g_l, \bm{p}^m_l, \bm{f}^p_l, \overline{\bm{f}^p_l})
\end{equation}
\begin{align}
    \overline{\bm{q}_l} = \frac{\overline{\bm{q}'_l}}{||\overline{\bm{q}'_l}||}
\end{align}
where $\bm{t}_l$ is the translation parameter of tooth $l$ while $\bm{q}_l$ is the rotation quaternion of it. $\omega^\text{Mo}$ is also a MLP block. Since the rotation quaternion should be a unit vector, we normalize the output $\overline{\bm{q}'_l}$.

When obtaining the motion $\overline{\bm{t}_l}, \overline{\bm{q}_l}$, we can calculate the predicted point clouds of each tooth as:
\begin{equation}
    \overline{\bm{P}_l} = \{\bm{R}_l \cdot (\bm{p} - \bm{p}^m_l) + \bm{p}^m_l + \bm{t}_l\}
\end{equation}
where $\bm{R}_l$ is the rotation matrix of quaternion $\bm{q}_l$ and we set the barycenter $\bm{p}^m_l$ of $\bm{P}_l$ as the rotation center for convenience.

\subsection{Feature Consistency Supervision}

\begin{figure}[tb]
    \centering
    \includegraphics[scale=0.23]{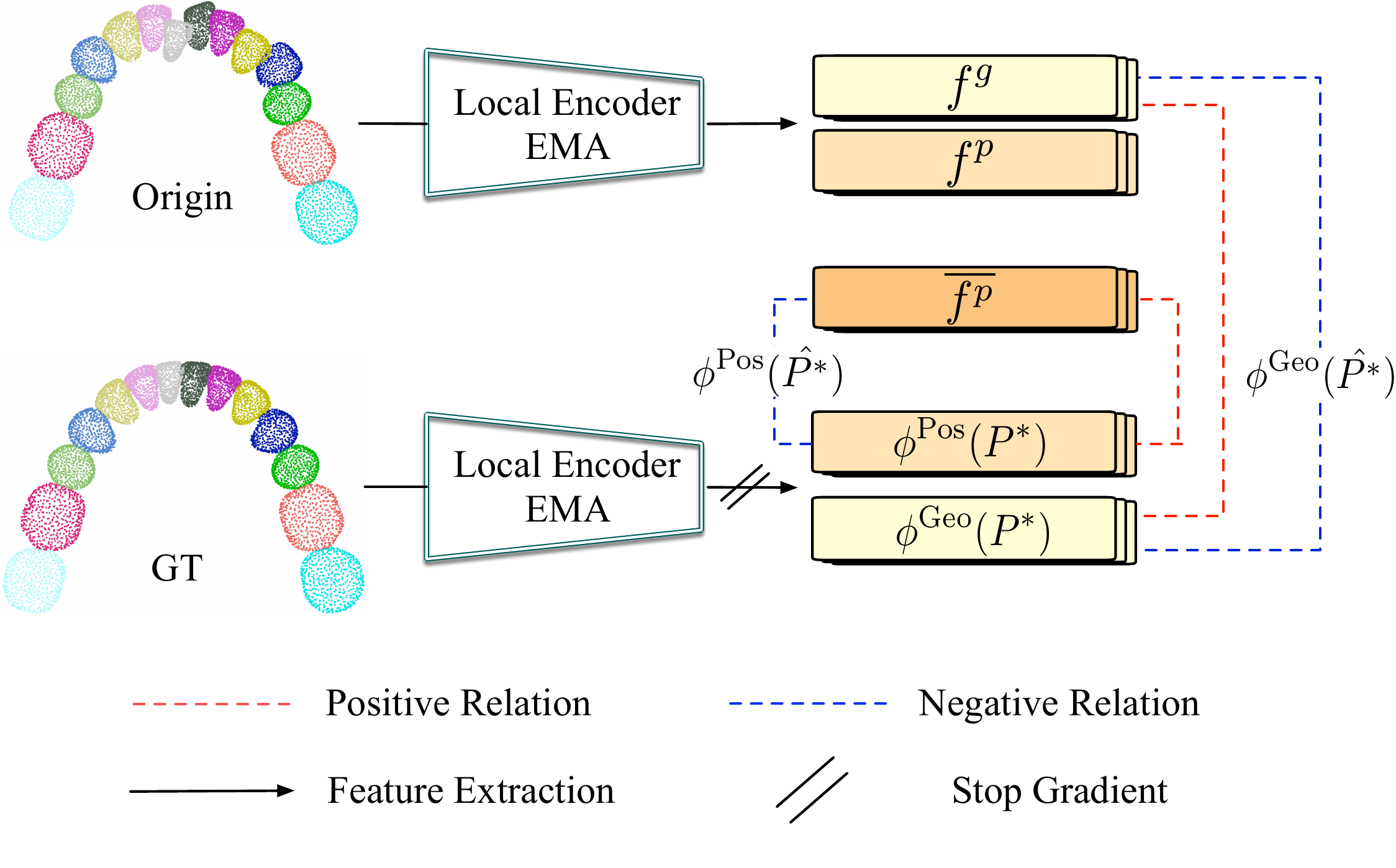}
    \caption{Overview of the Feature consistency supervision. $\bm{P^*}$ represents the point cloud of neat teeth from ground truth, and $\hat{\bm{P^*}}$ represents rearranged $\bm{P^*}$ with different categories.}
    \label{fig:feature_consistency_new}
\end{figure}

In order to force encoders to learn meaningful features of geometries and positions, we proposed feature consistency supervision. As the target positional features produced by the Feature Projection Module, they should be consistent with those extracted from the final dentition. The geometric features of the original dentition and target dentition should also be consistent. Inspired by recent advancements in contrastive learning\cite{BYOL, SimCLR}, we employ two identical sets of local encoders, mirroring the structure of those in the online network, to extract features from the original input and the ground truth, respectively. The parameters of these encoders are updated at each training iteration via the Exponential Moving Average (EMA) method using the online network’s parameters. We construct positive and negative pairs for two categories of features. The details are demonstrated in Figure \ref{fig:feature_consistency_new}. For geometric features, since the arrangement transformations are rigid, they can not change the geometric shape of teeth. Therefore, we use the same teeth before and after arrangement as positive pairs. Different categories of teeth, such as incisors and cuspids or cuspids and bicuspids, are considered negative pairs. For positional features, we consider the projected positional features $\overline{\bm{f}^p}$ and final positional features $\phi^{\text{Pos}}({\bm{P^*}})$ as positive pairs, while $\overline{\bm{f}^p}$ and the positional features of rearranged ground truth teeth $\phi^{\text{Pos}}(\hat{\bm{P^*}})$ are regarded as negative pairs. In DTAN, the feature consistency supervision can be formulated as follows:
\begin{equation}\label{feature consistency loss}
\begin{split}
\mathcal{L}_f =\frac{1}{4}(& 2 - \text{sim}(\bm{f}^g, \phi^{\text{Geo}}({\bm{P^*}})) + \text{sim}(\bm{f}^g, \phi^{\text{Geo}}(\hat{\bm{P^*}}))  + \\
&2 - \text{sim}({\overline{\bm{f}^p}, \phi^{\text{Pos}}(\bm{P^*}})) +  \text{sim}(\overline{\bm{f}^p}, \phi^{\text{Pos}}(\hat{\bm{P^*}})))
\end{split}
\end{equation}
where $\hat{\bm P^*}$ are teeth with different categories from ${\bm P^*}$, and \text{sim(·)} is the cosine similarity function, all these tensors have been normalized before \text{sim(·)}.

\subsection{Collision Supervision}

In real-world applications, some teeth need to be attached to another one. However, overlaps or gaps may exist in the results generated from neural networks, which are unacceptable. Based on this point, we need a differentiable function to constrain teeth tightly arranged so that our model can optimize it in the learning process, and it is better to be parallelizable to not cost much time in one learning iteration. Furthermore, teeth can be concave on the crown, so this function can not be convex-only for generalization. To the best of our knowledge, there is no such function for point cloud data. In our work, We propose an approximated collision loss function to satisfy all the requirements above.

\begin{figure}[tb]
    \centering
    \includegraphics[scale=0.18]{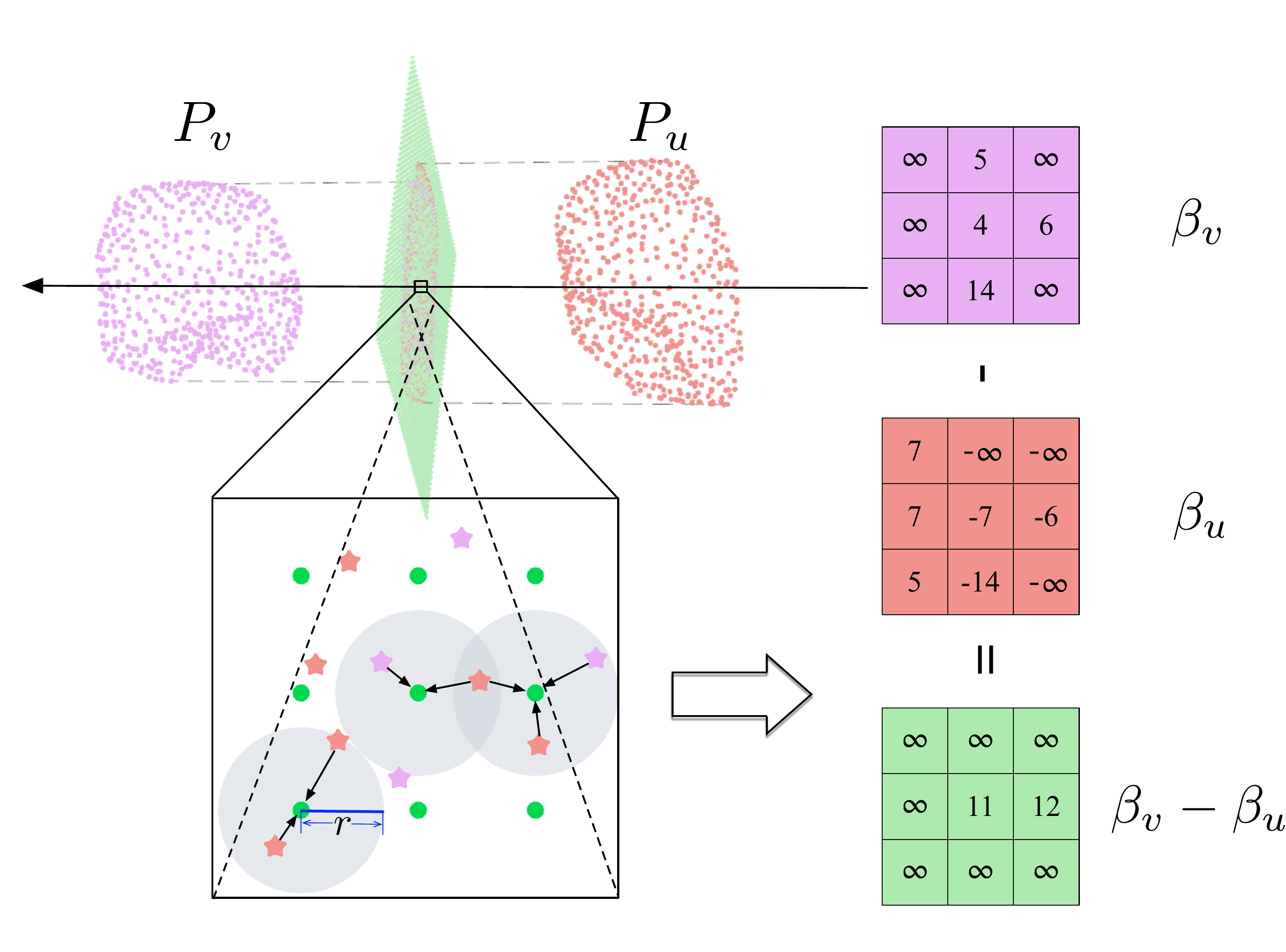}
    \caption{The calculation of Collision Supervision Function. First, we generate grid points (green points) in the middle vertical plane of two points clouds (red or purple points) and project two point clouds to the plane. Then, we use each grid to query the projected points close to it and obtain two depth maps $\bm{\beta_u}, \bm{\beta_v}$(where the gray circle indicates the query scope). At last, we get the overlap value or gap value by taking the minimal value of $\bm{\beta_v} - \bm{\beta_u}$, which is $11$ in this figure.}
    \label{fig:collision}
\end{figure}

First, we define the collision value between tooth $\bm{P}_u$ and tooth $\bm{P}_{v}$. As shown in Figure \ref{fig:collision}, The whole calculation process can be thought of as choosing a plane $\bm{\alpha}$ in three-dimensional space, projecting these two teeth to the plane to generate two depth distributions, and then calculating the overlapping and gapping values on the normal direction of plane $\bm{\alpha}$. Intuitively, we choose the middle vertical plane, which is perpendicular to the line of two mass centers $\bm{p}^m_u, \bm{p}^m_{v}$ and passing through the middle point $\bm{p}_{uv}$. Point clouds are in discrete data forms for three-dimensional shape models, so calculating the precise depth from two teeth is intractable. Therefore, we use an approximate method. Referring to the rendering process, we generate some grid points centered on point $\bm{p}_{uv}$ on $\bm{\alpha}$ with interval $R$. Along the normal direction of $\bm{\alpha}$, we calculate the directed distance between plane $\bm{\alpha}$ and each point of teeth $\bm{P}_u, \bm{P}_v$. Then we can get a maximal depth map $\bm{\beta}_u$ and a minimal depth map $\bm{\beta}_v$ as follows:
\begin{equation}
\begin{aligned}
    \bm{\beta}_u = \max_{\bm{g}}(-\infty, \max_{\bm{p}, \bm{p}' \in \mathcal{G}_{\bm{g}}}(||\bm{p} - \bm{p}'||_2)) \\
    \bm{\beta}_v = \min_{\bm{g}}(\infty, \min_{\bm{p}, \bm{p}' \in \mathcal{G}_{\bm{g}}}(||\bm{p} - \bm{p}'||_2))
\end{aligned}
\end{equation}
where $\bm{g}$ is the grid point, $\bm{p}$ is a point of $\bm{P}$, $\bm{p}'$ is the projected point of $\bm{p}$ and $\bm{p}'\in \mathcal{G}_{\bm{g}}$ indicates $\bm{p}'$ is in the query scope (gray circle with radius $r=R/\sqrt{2}$ in Figure \ref{fig:collision}) of grid point $\bm{g}$. In our implementation, we consider $\bm{p}' \in \mathcal{G}_{\bm{g}}$ if $||\bm{p}' - \bm{g}||_2 \leq r$. 

\begin{figure}[tb]
    \centering
    \includegraphics[scale=0.13]{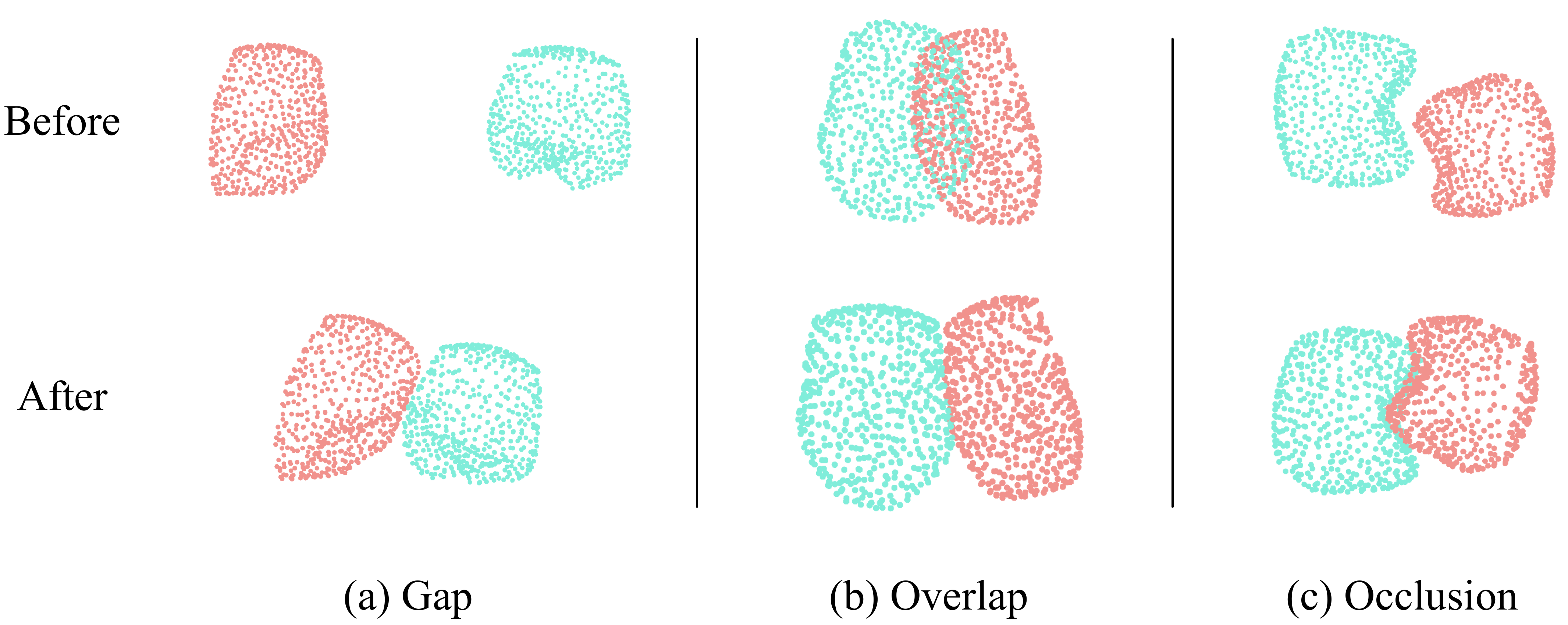}
    \caption{Validation of effectiveness of collision function in tiny teeth attachment tasks. In these examples, we initialize the positions of the two teeth as separate or embedded and calculate the collision function of the two teeth. Then, we use SGD\cite{bottou2010large} optimizer to translate the red tooth to be attached to the other one.}
    \label{fig:exp-collision-1}
\end{figure}
Comparing the values of corresponding grid points in two depth maps, the collision value $c_{uv}$ of these two teeth can be obtained by the minimal value of $\bm{\beta}_v - \bm{\beta}_u$. $c_{uv} < 0$ indicates $\bm{P}_u$ and $\bm{P}_v$ are collided with each other while $c_{uv} > 0$ indicates there is gap exists between these two teeth. The formulation of the collision value is as follows:
\begin{equation}
    c_{uv}=\min_g\{\bm{\beta}_v^g - \bm{\beta}_u^g\}.
\end{equation}

With the collision value defined, we can constrain two teeth to be attached using the following formula:
\begin{equation}
    \mathcal{L}_c^{uv} = c_{uv}^2.
\end{equation}

We present three examples of collision supervision in Figure \ref{fig:exp-collision-1}. In these examples, we set two teeth at their initial positions at first and then translate one tooth, which is red, to get it attached to another one. We use gradient descent to optimize the position of the red tooth iteratively, and the final results verify the effectiveness. More details can be found in Section \ref{subsubsec:clf}.

\begin{figure}[tb]
  \centering
    \includegraphics[width=0.6\linewidth]{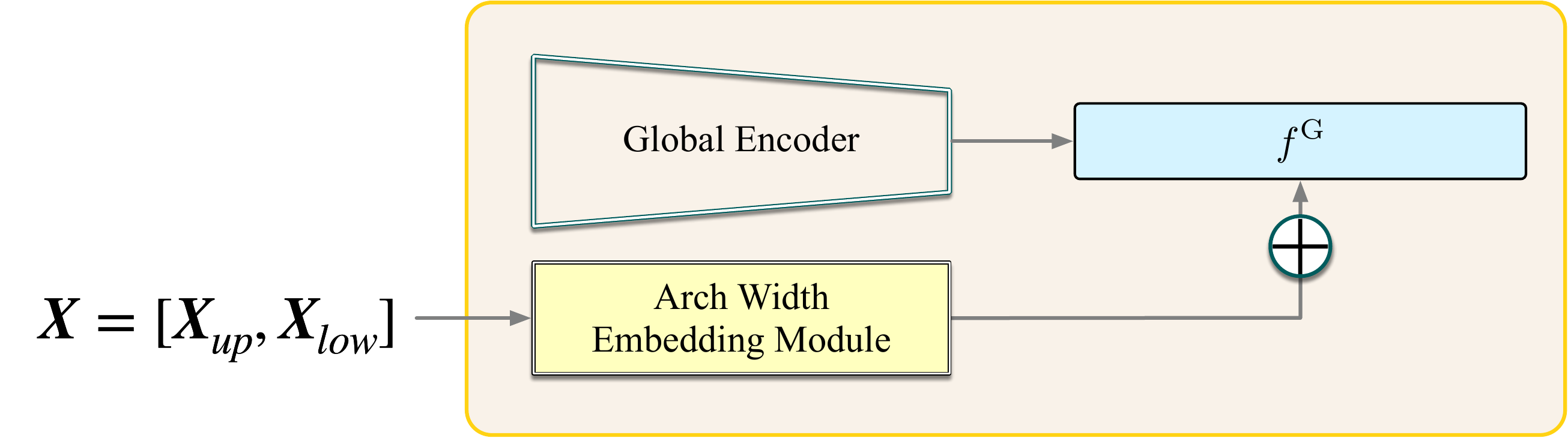}
    \label{fig:sub1}
    \includegraphics[width=0.6\linewidth]{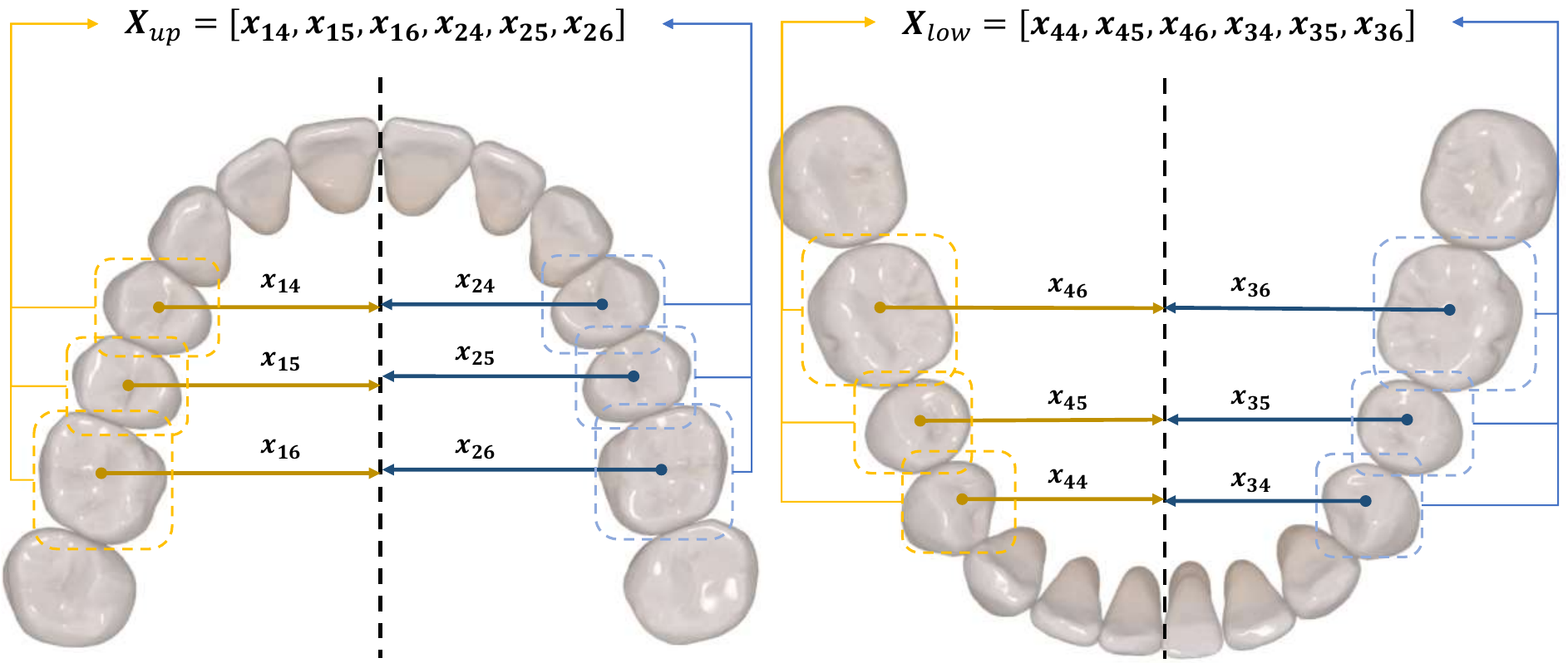}
    \label{fig:sub2}
  
  \caption{The workflow of the conditional prediction generation module using the arch widths. We use $\bm{X}_{up}$ and $\bm{X}_{low}$ to represent the arch widths of the upper and lower jaws, respectively, We extract the features through the Arch Width Embedding Module, and add them to the global features to control the network's predictions.}
  \label{fig:arch width}
\end{figure}

\subsection{Conditional Prediction Generation}
With DTAN, we are able to obtain overall satisfactory tooth arrangement results. However, doctors often have different design preferences for tooth arrangement plans in clinical practice. In addition, based on the patient’s demands and factors such as his alveolar bone and facial shape, the doctor will also customize a tooth arrangement plan, mainly reflected in the design of the dental arch width. Therefore, a clinically usable network should be able to generate diverse tooth arrangement results guided by simple conditions, and the results should comply with orthodontic rules.

In the previous works, TANet\cite{wei2020tanet} append a random vector $\xi \in \mathbb N(0,\text{I})$ to the input features in training to generate diverse predictions. Although this solution is easy to implement, it cannot produce reasonable results according to the needs of doctors and patients. To this end, in our work, we propose a simple conditional tooth arrangement result generation network (named \textbf {C-DTAN}). Doctors only need to input a simple vector of the target dental arch widths as a prompt, and the C-DTAN will be able to make corresponding predictions and ensure that the dentition is reasonable and satisfactory. Compared with DTAN, it requires only slight modifications to the network structure and adds almost no computational cost.

The workflow of the conditional prediction generation module is shown in Figure \ref{fig:tooth_labels}. Building upon the vanilla DTAN, we integrated a dental arch width embedding module (consists of two linear layers, utilizing leaky ReLU as the activation function in our implementation) to enhance the input vector $\boldsymbol {X}$, without making any other changes to DTAN. Drawing from orthodontic expertise, we utilize $\boldsymbol {X}=[\boldsymbol{X}_{up}, \boldsymbol{X}_{low}]$, a vector consisting of the directional distances from the masses of the first and second premolars, as well as the first molars of the upper and lower jaws, to the midsagittal plane to simplify the representation of dental arch widths. These directional distances serve as inputs to the embedding module. Subsequently, we add the output embeddings to the global features, enabling the network to perceive the target widths of the dental arch. The value of $\boldsymbol{X}$ is calculated through ground truth during training and can be given by experienced experts or doctors during inference.

\subsection{Loss Function}
In this section, we introduce our loss function. Symbols with overline indicate the predictions and superscript * indicates the ground truth.

\paragraph{Reconstruct Loss}
In our work, we choose MSE Loss as reconstruct loss function instead of chamfer distance loss to enhance the learning of teeth rotations. We minimize the distances between each corresponding point in the prediction and the ground truth:
\begin{equation}
    \mathcal{L}_r = \frac{1}{|\bm{L|}|\overline{\bm{P}_l}|}\sum_{l\in \bm{L}}\sum_{\overline{\bm{p}} \in \overline{\bm{P}_l}}||\overline{\bm{p}} - \bm{p}^*||_2.
\end{equation}

\paragraph{Transformation Parameter Loss}
To enhance the supervision of teeth rotations, we also use the transformation parameter, which contains rotation quaternion, to supervise the network directly:
\begin{equation}
    \mathcal{L}_p = \frac{1}{|\bm{L}|}\sum_{l \in \bm{L}}||\overline{\bm{q}_l} - \bm{q}_l^*||
\end{equation}
where $\overline{\bm{q}}_l$ and $ \bm{q}_l^*$ indicate the predicted quaternion and ground truth quaternion of tooth $l$, respectively.

\paragraph{Feature Consistency Loss}
We apply the feature consistency loss as Eq.\ref{feature consistency loss}.

\paragraph{Collision Supervision Loss}
In tooth arrangement, we expect the adjacent teeth in the same jaw and teeth with occlusal corresponding teeth in the opposite jaw to be attached. So, the collision supervision loss can be formulated as follows:
\begin{equation}
    \mathcal{L}_c = \frac{1}{|\bm{L}||\mathcal{N}|}\sum_{u\in \bm{L}}\sum_{v \in \mathcal{N}} \mathcal{L}_c^{uv}
\end{equation}
where $\mathcal{N}(u) = \text{NBR}(u) \cup \text{OPS}(u)$ means the tooth in neighbor or the opposite tooth in the other jaw.

Finally, the total training loss is obtained by combining all these losses together with hyperparameters.
\begin{equation}
    \mathcal{L} = \lambda_r \mathcal{L}_r +  \lambda_p\mathcal{L}_p + \lambda_f \mathcal{L}_f + \lambda_c \mathcal{L}_c.
\end{equation}

\subsection{Augmentation}
In orthodontic clear aligner treatment planning, a case has many transitional statuses called stages, all of which aim toward the target poses. This indicates that most situations of one case should have the same target. Since there are often only a few data in this task, performing data augmentation for existing data is crucial.
In our work, we first apply random transformations to the inputs or targets to generate diverse inputs. Following the previous works, all individual teeth are randomly rotated by an angle within $[-30^{\circ},+30^{\circ}]$, in a random direction and translated by a distance vector from the Gaussian distribution $\mathbb N(0, 1^2)$ of unit $\text{mm}$. However, in most situations, these augmentation methods generate inputs near the original and target poses, ignoring the transitional status. Therefore, in addition to naively random augmentation, we simulate the staging process by interpolating the intermediate process between the original and target poses.

\section{Experiments}

\subsection{Datasets} 
In order to verify the effectiveness of our method, we collected 909 cases from real-world orthodontic treatment plans. Each case has its ground truth transformation parameters given by experts, and has been segmented, completed, and labeled with its category $l \in \bm{L}$, where $\bm{L} = \{11-17, 21-27, 31-37, 41-47\}$ is shown in Figure \ref{fig:tooth_labels}. However, some of these cases have overlaps, gaps, or asymmetries due to operational mistakes. Therefore, we filter a batch of higher-quality data from them to form a new dataset (i.e., the High-Quality Dataset). For network training, we split the two datasets randomly into three sets. We use 382, 54, and 108 cases as training, validation, and test sets for the High-Quality Dataset, while 636, 91, and 182 cases as training, validation, and test sets for the Full Dataset.

Before arrangement, the previous works\cite{wei2020tanet,wang2022tooth} need to register the whole original dentition and ground truth dentition by iterative closest point (ICP)\cite{besl1992method} registration method. Then, they calculate the ground truth transformations by ICP tooth by tooth. This method introduces system errors and eliminates overall translations. The previous works must include this step because they only have pairs of original and final dentition scanning models. Unlike them, our dataset already contains the accurate transformation parameters of arrangement. For normalization, we transform the coordinates system to ensure all cases are in a uniform direction, and at the same time, the coordinate origin sits at the center of the teeth $\{11-41\}$.

\begin{table*}[tb]
    \caption{Quantitative comparison of different methods. The unit of {\upshape $\text{ME}_\text{point}$} and {\upshape $\text{ME}_\text{trans}$} is {\upshape mm} while that of {\upshape $\text{ME}_\text{rotat}$} is degree. The {\upshape LandmarkNet} is the network from Wang's work\cite{wang2022tooth} and the symbol $^ {\dag}$ indicates {\upshape LandmarkNet} without its landmarks. More related details can be found in table \ref{tab:landmark_results}.}
    \label{tab:overal_results}
    \centering
    \begin{tabular}{lrrrrrrrrr}
        \toprule
        \multirow{2}{*}{Method}  &  \multicolumn{4}{c}{High-Quality Dataset} & & \multicolumn{4}{c}{Full Dataset} \\ 
                & $\text{ME}_\text{point}$$\downarrow$ & $\text{ME}_\text{trans}$$\downarrow$ &  $\text{ME}_\text{rotat}$$\downarrow$ & $\text{AUC}$$\uparrow$ & & $\text{ME}_\text{point}$$\downarrow$ & $\text{ME}_\text{trans}$$\downarrow$ & $\text{ME}_\text{rotat}$$\downarrow$ & $\text{AUC}$$\uparrow$ \\
        \midrule
        Before Arrangement*  & 3.0594 & 2.9603 & 5.5252 & 33.59 & & 3.2907 & 3.2015 & 5.0268 & 29.65 \\
        PointNet \cite{qi2017pointnet}        & 1.2118 & 1.1137 & 3.4915 & 67.52 & & 1.4576 & 1.3837 & 3.1109 & 63.53 \\
        PointNet++ \cite{qi2017pointnet++}      & 1.1900 & 1.1134 & 3.0723 & 67.89 & & 1.5111 & 1.4493 & 2.9046 & 62.92 \\
        DGCNN \cite{wang2019dynamic}           & 1.1692 & 1.0827 & 3.2393 & 68.50 & & 1.5343 & 1.4677 & 3.0217 & 61.93 \\
        PointNext \cite{qian2022pointnext}       & 1.2138 & 1.1271 & 3.2724 & 67.41 & & 1.5529 & 1.4869 & 3.1009 & 60.76\\
        PSTN \cite{li2020malocclusion}            & 1.1229 & 1.0466 & 2.9727 & 70.06 & & 1.4294 & 1.3595 & 3.0194 & 64.30 \\
        TAligNet \cite{lingchen2020iorthopredictor}        & 1.0328 & 0.8717 & 4.3021 & 72.80 & & 1.1224 & 0.9819 & 3.9496 & 69.24 \\
        TANet \cite{wei2020tanet}           & 0.9462 & 0.8520 & 3.0668 & 75.96 & & 0.9523 & 0.8712 & 2.7471 & 74.95 \\
        $\text{LandmarkNet}^{\dag}$ \cite{wang2022tooth} & 0.9298 & 0.8367 & 3.0569 &  76.49 & & 0.9578& 0.8688 & 2.7421& 74.94 \\
        DTAN (ours)            & \bfseries{0.7952} & \bfseries{0.7125} & \bfseries{2.6377} & \bfseries{81.48} & & \bfseries{0.9088} & \bfseries{0.8279} & \bfseries{2.6387} & \bfseries{78.18} \\
        \midrule
        C-DTAN (ours) & \bfseries{0.7265} & \bfseries{0.6407} & \bfseries{2.5348} & \bfseries{83.99} & & - & - & - & - \\
        \bottomrule
    \end{tabular}

\end{table*}

\subsection{Training Details and Evaluation Metrics}
We sample $N = 512$ points from the point cloud data of each tooth using the Farthest Point Sampling method. We utilize PointNet without STN as our encoder. Both the Feature Projection Module and the Motion Regression Module are 3-layer MLP. Meanwhile, $\bm{f}^g_l, \bm{f}^p_l, \bm{h}^g_l, \bm{h}^p_l, \overline{\bm{f}^p_l}$ vectors are 512-dimensional, and $\bm{f}^{\text{Global}}$ vectors are 1024-dimensional. The model is implemented with PyTorch, and we trained it for $1000$ epochs with batch size setting as $16$, using an NVIDIA RTX 3090 GPU with 24 GB memory. We use SGD\cite{bottou2010large} as the optimizer during the learning process, with an initial learning rate of $1 \times 10^{-3}$ and a weight decay of $1 \times 10^{-4}$. In collision calculation, we set interval $R=0.3\text{mm}$ and the resolution of grids are set as $50 \times 50$. Regarding the value of hyperparameters, we set $\lambda_r=0.5, \lambda_p=20.0, \lambda_f=1.0, \lambda_c=2.0$.

To ensure a fair comparison with the previous works, we use metrics similar to TANet \cite{wei2020tanet}. We calculate the PCT@$K$ metric using the prediction error percentage more minor than the threshold $K$. We use mean point-wise error as our predicting error between the predicted point cloud and ground truth to get the PCT curve with $\max K$ to be $3 \text{mm}$ and interval to be $0.01\text{mm}$. The metric $\text{AUC}$ indicates the area under this PCT curve. The mean point-wise distance error $\text{ME}_\text{point}$ is also calculated. In addition to evaluating the accuracy of transformation parameters, we use mean translates error and mean rotation error record as $\text{ME}_\text{trans}$ and $\text{ME}_\text{rotat}$.

\begin{figure*}[tb]
    \centering
    \includegraphics[scale=0.5]{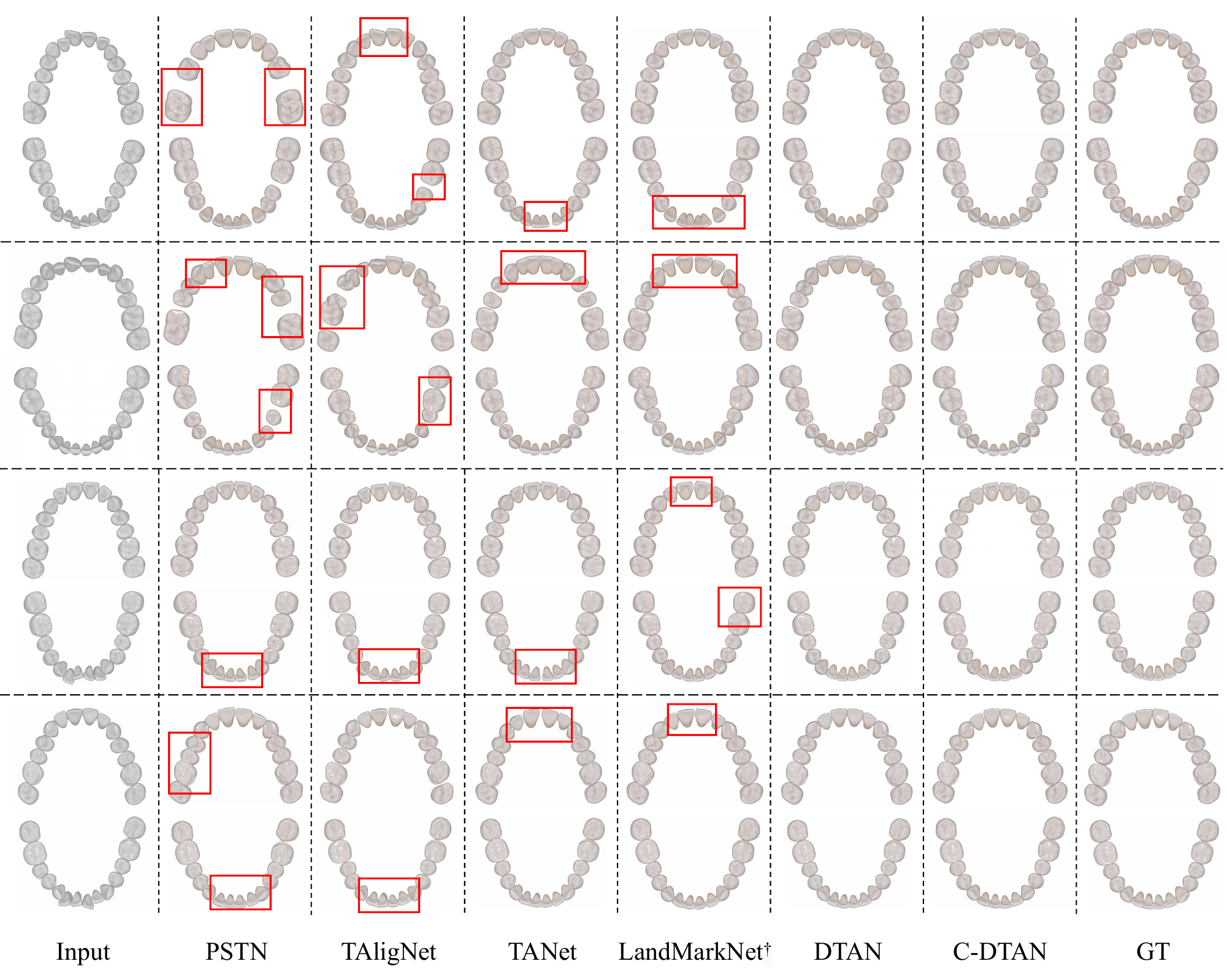}
    \caption{Visualization results of different methods. The red boxes in the figure indicate where the predictions of other methods are unreasonable. In these three cases, our DTAN and C-DTAN generate the prettiest while keeping the spaces of adjacent teeth within a reasonable range.}
    \label{fig:visulization_results}
\end{figure*}

\begin{figure*}[!ht]
    \centering
    \includegraphics[scale=0.15]{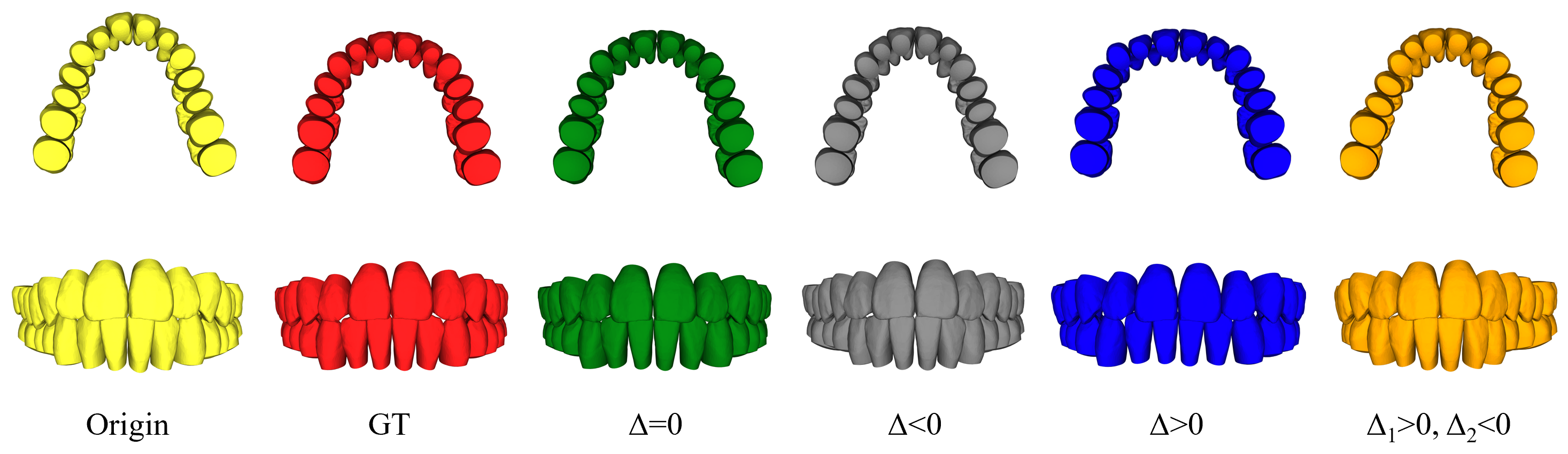}
    \caption{Visualization of conditional predictions generated by C-DTAN by different arch width offsets $\Delta$. $\Delta=0$ means using the original X as the arch width input, $\Delta < 0$ and $\Delta > 0$ represent the results of arch contraction and arch expansion operations, respectively. And the last result (orange) represents the result of an asymmetric dental arch.}
    \label{fig:arch_width_vis}
\end{figure*}

\begin{table}[h]
    \caption{Quantitative comparison on the dataset with landmarks.}
    \label{tab:landmark_results}
    \centering
    \begin{tabular}{ccccc}
        \toprule
        Method & $\text{ME}_\text{point}\downarrow$ & $\text{ME}_\text{trans}\downarrow$ & $\text{ME}_\text{rotat}\downarrow$ & $\text{AUC}\uparrow$ \\
        \midrule
        Before Arrangement* & 3.1310 & 3.0290 & 5.3182 & 30.55 \\
        TANet \cite{wei2020tanet} & 0.9717 & 0.8822 & 3.0743 & 74.83 \\
        LandmarkNet \cite{wang2022tooth} & 0.9076 & 0.8368 & 2.7831 & 77.08 \\
        DTAN(ours) & \bfseries 0.8690 & \bfseries 0.7880 & \bfseries 2.7481 & \bfseries 78.55\\
        \bottomrule
    \end{tabular}
\end{table}

\subsection{Comparison with Other Methods}
To validate the effectiveness of our method, we compared it with some basic point cloud methods as well as some previous learning-based automatic tooth arrangement methods. The quantitative results of all methods are shown in Table \ref{tab:overal_results}. In all of the evaluation metrics, our method achieves the best performance on both the High-Quality dataset and the full dataset. The result could be partially uncontrollable since there are some defective cases in the full dataset. Figure \ref{fig:visulization_results} shows the visualization results, which also confirm the superiority of our method. In addition, we test our method's inference time, which decreases that of TANet \cite{wei2020tanet} from $48\text{ms}$ to $21\text{ms}$. More results, including user study, are displayed in the section \ref{supplemental results}.

In order to verify the capabilities of our proposed C-DTAN, we design a set of comparative visualization experiments. Specifically, we offset the input dental arch width vector $\boldsymbol{X}$, where the offset is given by $\Delta = [\Delta_1, \Delta_2]$. Among them, \(\Delta_1\) is applied to the vector values represented by the six teeth on the left, i.e., $x_{\{14,15,16,44,45,46\}}$; \(\Delta_2\) is applied to the remaining right tooth vectors. By controlling the value of \(\Delta\), we can let the network predict the results of arch contraction and expansion, or even asymmetric arches. This ability is difficult to achieve using conventional training methods due to the lack of diverse training data for different situations. The visualization results are shown in the Figure \ref{fig:arch_width_vis}.

Wang \textit{et al.}\cite{wang2022tooth} proposed a method, which we call LandmarkNet in this paper, using geometric landmarks of teeth to constrain the arrangement results. However, acquiring precise landmarks data requires orthodontic experts and the unbearable cost of their time. Therefore, we compare their method by first using their network without landmarks in our High-Quality Dataset and Full Dataset, which are presented in Table \ref{tab:overal_results}. To further validate the results, we construct a tiny dataset with landmarks labeled by experts. The dataset consists of 319 cases, the same as LandmarkNet, with landmarks also consistent with it. The results show that our method even outperforms LandmarkNet without using any landmarks. We attribute it to the complex loss function of LandmarkNet that may harm the optimization. Meanwhile, our collision supervision can substitute some landmark constraints in a more uniform way.

\begin{table}[h]
    \caption{Ablation study on High-Quality dataset.}
    \label{tab:ablation study}
    \centering
    \begin{tabular}{ccccrr}
        \toprule
        Proj & Prop & Col & $\text{Aug}^{\dag}$ & $\text{ME}_\text{points}\downarrow$ & $\text{AUC}\uparrow$ \\
        \midrule
          &  &  &  & 0.9578 & 75.52 \\
         \checkmark &  &  &  & 0.9256 & 76.71 \\
         \checkmark & \checkmark & &  & 0.8869 & 78.12 \\
         \checkmark & \checkmark & \checkmark &  & 0.8421 & 79.75 \\
          & \checkmark & \checkmark & \checkmark & 0.8194 & 80.59  \\
         \checkmark &  & \checkmark & \checkmark & 0.8658 & 78.94 \\
         \checkmark & \checkmark &  & \checkmark & 0.8325 & 79.97  \\
         \checkmark & \checkmark & \checkmark & \checkmark & \bfseries{0.8071} & \bfseries {81.48} \\
        \bottomrule
    \end{tabular}

\end{table}

\subsection{Ablation Study}

\subsubsection{The Effectiveness of Each Module}
We remove the Feature Propagation Module (Prop), the Feature Projection Module (Proj), the collision loss (Col), and interpolation-based data augmentation ($\text{Aug}^{\dag}$) from DTAN as our baseline. We use global features, local features, and barycenters as input to the MLP decoder and ensured that the total number of feature channels remained constant with the same data augmentation as TANet. Table \ref{tab:ablation study} shows the importance of each sub-module on the High-Quality Dataset. With the Feature Propagation Module and the Feature Projection Module, DTAN can better learn teeth features and understand the correlations between teeth in a decoupled way. Interpolation-based data augmentation effectively improves the diversity as well as rationalization of the data. The collision loss significantly alleviates the non-conformity of orthodontic rules in the predicted dentition and is also helpful for further error reduction.

\subsubsection{Collision Supervision Function}
\label{subsubsec:clf}

\begin{figure*}[tb]
    \centering

        \includegraphics[scale=0.26]{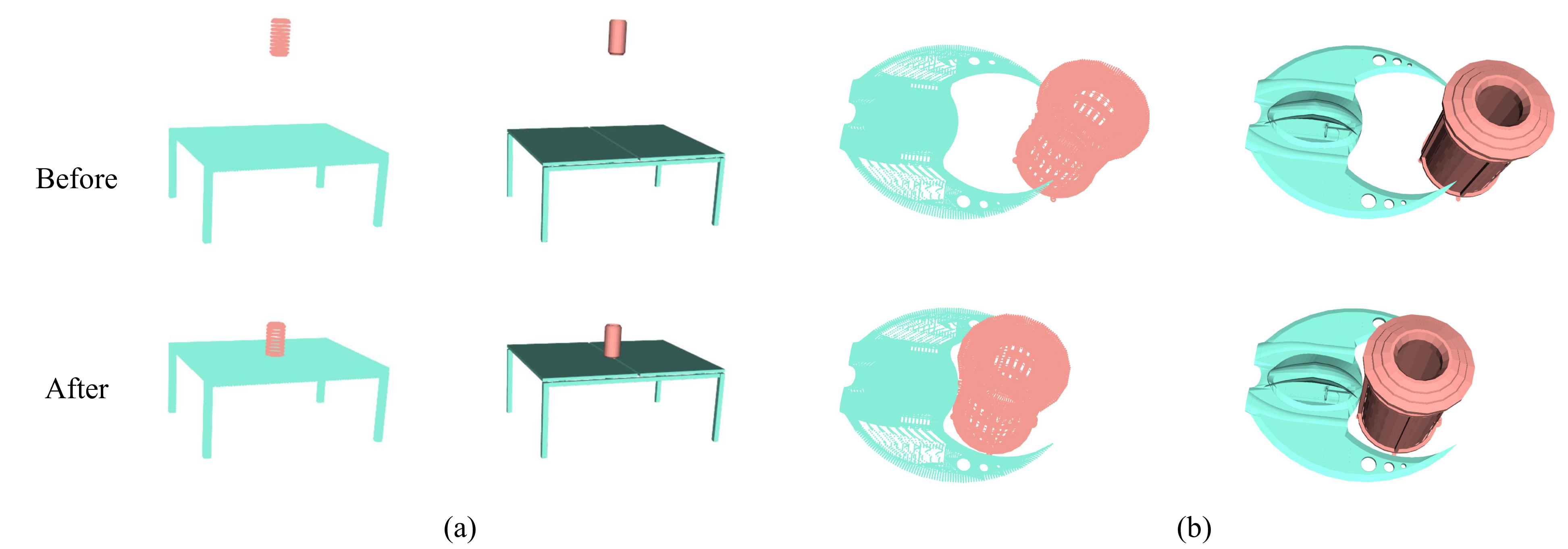}
    \caption{Validation of effectiveness of collision function in ShapeNet\cite{shapenet2015} objects attachment tasks. In each case, the left side is point cloud representation, while the right side is mesh representation. The experimental setup is as the same as Figure \ref{fig:exp-collision-1}.}
    \label{fig:exp-collision-2}
\end{figure*}


\begin{table}[tb]
    \caption{Comparisons for running time of Collision Supervision Function in a training iteration.}
    \label{tab:running_time}
    \centering
    \begin{tabular}{lr}
         \toprule
         Process & Time (ms) \\
         \midrule
         Forward & 97 \\
         Loss Calculation & 33 \\
         Backward & 74 \\
         Collision Function Calculation & 19 \\
         \bottomrule
    \end{tabular}

\end{table}

In order to visualize the effectiveness and versatility of the collision supervision function, we test this function on some tiny tasks shown in Figure \ref{fig:exp-collision-1}. For different initial situations on the left side, we iteratively predict translation parameters for blue teeth with the SGD\cite{bottou2010large} optimizer to eliminate the gap or overlap of teeth. Moreover, we verify collision supervision in ShapeNet\cite{shapenet2015} which is presented in Figure\ref{fig:exp-collision-2} with the same experimental setup. The results show the practicality of collision supervision. We also evaluate the actual effect on tooth arrangement with the results of reducing the frequency of overlaps or gaps from 3.88 to 1.31 on average.
Notably, it can even constrain the occlusal relation between upper and lower teeth. Referring to Table \ref{tab:running_time}, our collision supervision only costs $19\text{ms}$ in a training iteration, which is reasonably fast. 

\begin{figure}[tb]
    \centering
    \includegraphics[scale=0.1]{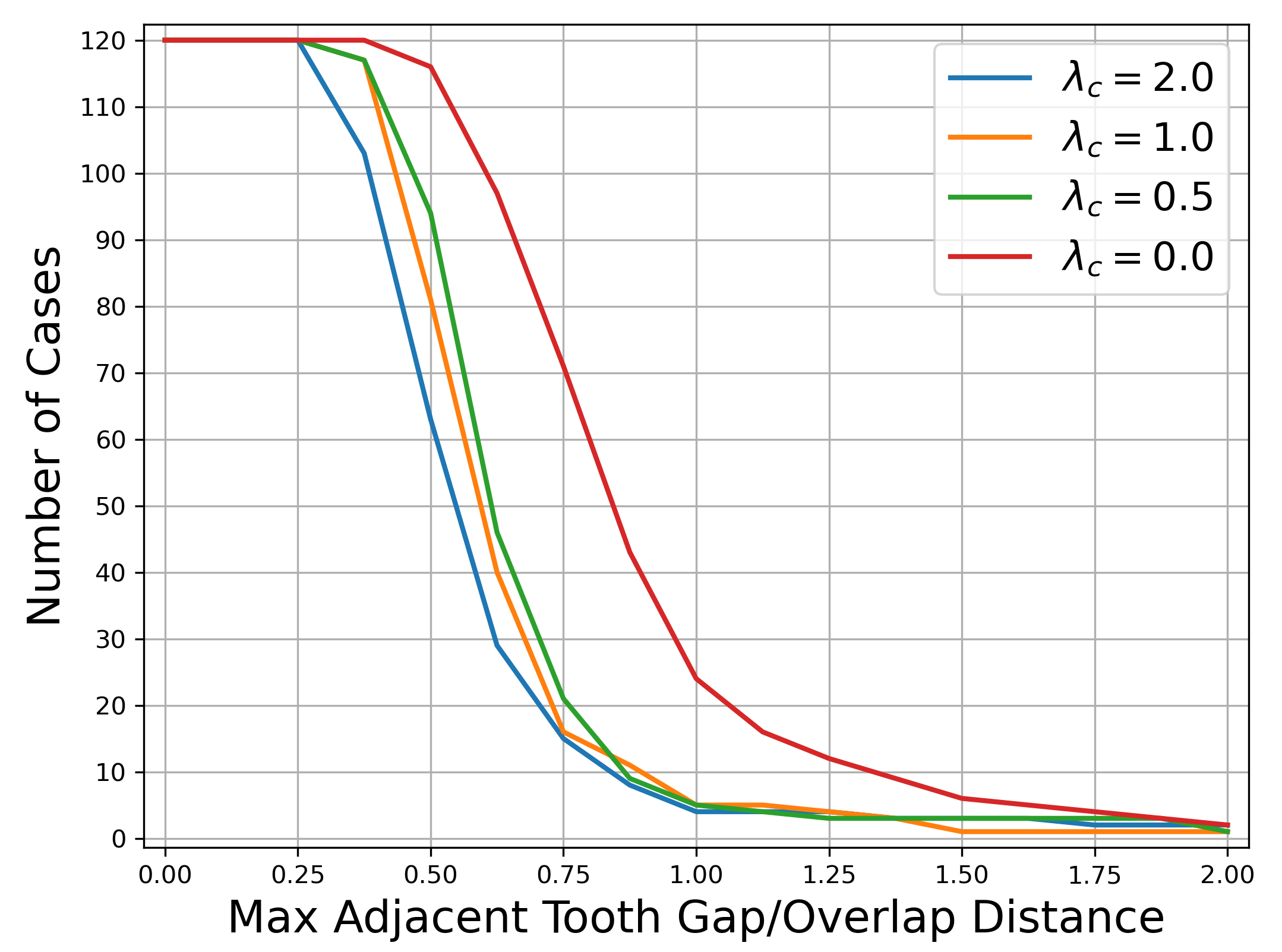}
    \caption{Number of cases whose maximum gap or overlap distances between adjacent teeth id larger than the threshold.}
    \label{fig:collision-evalution}
\end{figure}

In order to show the effectiveness of collision supervision more comprehensively, we further measure the gaps and overlaps of predictions, shown in Figure \ref{fig:collision-evalution} and Table \ref{tab:collision}. To accurately evaluate the gaps or overlaps between adjacent teeth, we calculate the signed distance for each pair of adjacent teeth with their meshes. For one case, we use mean absolute signed distance $\text{mean} |d|$, maximum absolute signed distance $\max |d|$, and the number of pairs with overlap or gap larger than $0.5\text{mm}$. Table \ref{tab:collision} shows the average indices of all testing cases. By using collision supervision, the number of gaps or overlaps can be reduced from $3.88$ to $0.94$ on average. At the same time, Figure \ref{fig:collision-evalution} visualizes the number of cases whose maximum absolute signed distance between adjacent teeth is higher than the threshold. Our collision supervision is of significant assistance in eliminating the gaps or overlaps between adjacent teeth.

\begin{table}
    \caption{Comparisons for collision test of different $\lambda_a$ weights.}
    \label{tab:collision}
    \centering
    \begin{tabular}{lrrr}
        \toprule
        $\lambda_c$ & $|d|_\text{mean}$($\text{mm}$) & $|d|_\text{max}$($\text{mm}$) & $\overline{N}_{|d| > 0.5}$ \\
        \midrule
        $0$     & 0.279 & 0.866 & 3.88 \\
        $0.5$   & 0.206 & 0.641 & 1.59 \\
        $1.0$   & 0.196 & 0.612 & 1.31 \\
        $2.0$   & 0.179 & 0.574 & 0.94 \\
        \bottomrule
    \end{tabular}
\end{table}

\begin{figure}[tb]
    \centering
    \includegraphics[scale=0.4]{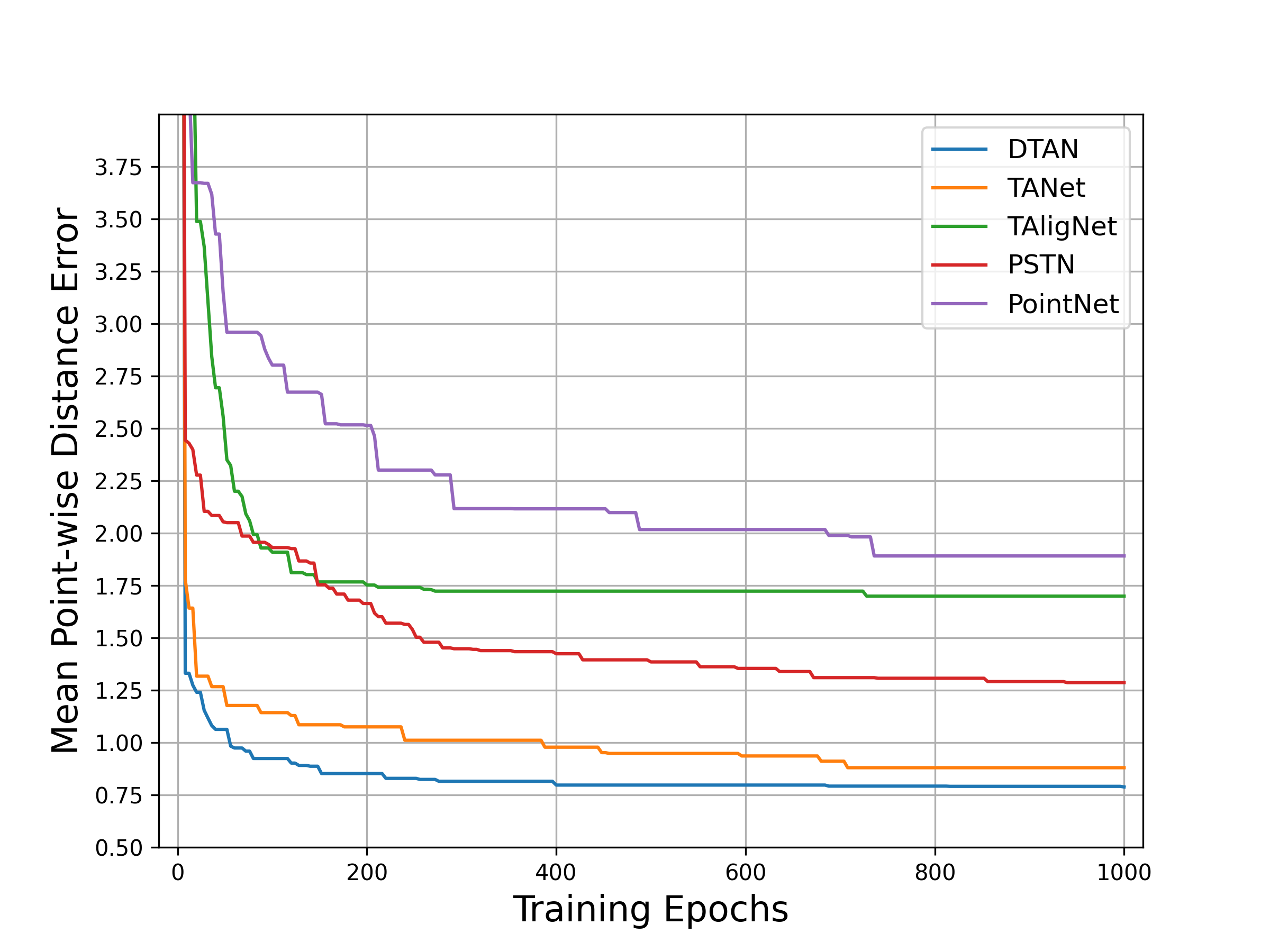}
    \caption{Visualization of mean point-wise distance error changing with training epochs.}
    \label{fig:convergence}
\end{figure}

\subsection{Supplemental Results}\label{supplemental results}
We plot the mean point-wise distance error changes of different methods on the validation set with the number of training epochs in Figure \ref{fig:convergence}, demonstrating the efficiency of our DTAN's training process. Comparing the previous methods, DTAN has a faster convergence speed and only requires a small number of iterations to achieve a small test error. In particular, DTAN only needs about 400 epochs to reach almost optimal accuracy. This may be because we decouple features and tasks, making learning easier for the model. In addition, using the transform layers for autocorrelated feature propagation can converge faster than the previous method based on GRU \cite{wei2020tanet}.

\subsection{User Study}

\begin{table}
    \caption{Comparisons for the mean scoring results from experts. For the four questions Q1 to Q4, Each case is scored from 0 to 5 point, where 5 means the best.}
    \label{tab:user_study}
    \centering
    \begin{tabular}{p{4cm}rrrr}
        \toprule
        Method & Q1 & Q2 & Q3 & Q4 \\
        \midrule
        TAligNet \cite{lingchen2020iorthopredictor} & 1.69 & 2.42 & 1.26 & 1.52 \\
        TANet \cite{wei2020tanet} & 3.53 & 3.89 & 3.06 & 3.26 \\
        Ground Truth   & 3.87 & 3.98 & \bfseries{4.05} & 3.85 \\
        DTAN (ours)   & \bfseries{4.04} & \bfseries{4.21} & 3.81 & \bfseries{3.92} \\
        \bottomrule
    \end{tabular}

\end{table}

We also validate the effectiveness of our model subjectively by a professional orthodontic expert. We invite experts to design four aspects to evaluate where a tooth arrangement result is satisfying. These four aspects are:
\begin{itemize}
    \item Q1: Whether the arrangement of single jaw meets the needs of orthodontic treatment?
    \item Q2: Whether the occlusal relationship between upper jaw and lower jaw meets the needs of orthodontic treatment?
    \item Q3: Whether the relationships between adjacent teeth meet the needs of orthodontic treatment? 
    \item Q4: Whether you are satisfied with the arrangement? 
\end{itemize}
We ask $3$ experts to score these four questions from $0$ to $5$, where $5$ indicates the best. For each case, we compare the results of TANet \cite{wei2020tanet}, TAligNet \cite{lingchen2020iorthopredictor}, ground truth, and our DTAN. We randomly shuffle these results to get fair comparisons. $54$ cases, randomly choosing from test set, are scored in total, and the results are shown in Table \ref{tab:user_study}. In general, our method gets competitive results even compared with experts. For the arrangement of the single jaw and the occlusal relationship, DTAN outperforms others and gets scores above $4$ on average. However, our DTAN still has some defects, e.g., the issue of gaps and overlaps is still not resolved entirely.

\section{Conclusion}
In this paper, we propose DTAN, a differentiable collision-supervised network for tooth arrangement. With a decoupling perspective, DTAN decomposes the arrangement task into target pose perception and assisted transformation regression tasks. And DTAN exploits the consistency of teeth to benefit feature learning with hidden features decoupled. In addition, a novel collision loss is proposed to reduce the possible overlaps or gaps between teeth of predicted dentition, which can be applied to other point cloud tasks. In the three datasets we constructed, DTAN achieves the best performance compared with other existing methods. Furthermore, we propose an arch-width guided tooth arrangement network, named C-DTAN, to control the arrangement results with arch width. The ablation study and the other experiments demonstrate the effectiveness of each sub-module of our methods. 

Although we achieved an exciting result for tooth arrangement, there is still some future work. For example, the results of tooth arrangement could vary according to patients' roots and alveolar bones. A reasonable approach is to use information such as CBCT data to plan tooth arrangement results and avoid collisions. Therefore, using absolute similarity between ground truth and predictions as the only evaluation metric is not comprehensive. New evaluation metrics that could measure tidiness and functionality are needed.

\bibliographystyle{plain}
\bibliography{references}  






\end{document}